\pdfoutput=1

\documentclass[11pt]{article}

\usepackage[preprint]{acl}

\usepackage{times}
\usepackage{latexsym}

\usepackage{comment}
\usepackage{booktabs}
\usepackage{multirow}
\usepackage{url}

\usepackage[T1]{fontenc}

\usepackage[utf8]{inputenc}

\usepackage{microtype}

\usepackage{inconsolata}

\usepackage{amssymb,amsmath}
\usepackage{xcolor}
\usepackage{pythonhighlight}
\usepackage{multicol}
\lstnewenvironment{mypython}[1][]{\lstset{style=mypython,#1}}{}

\usepackage{algorithm}
\usepackage{algpseudocode}
\usepackage{amsmath}
\usepackage{tcolorbox}

\title{\texttt{WiDe-analysis}: Enabling One-click Content Moderation Analysis on Wikipedia's Articles for Deletion}

\author{
\textbf{Hsuvas Borkakoty\textsuperscript{1}},
\textbf{Luis Espinosa-Anke\textsuperscript{1,2}}
\\
\textsuperscript{1}Cardiff NLP, School of Computer Science and Informatics, Cardiff University, UK 
\\
\textsuperscript{2}AMPLYFI, UK
\\
\small{
   \texttt{\{borkakotyh,espinosaankel\}@cardiff.ac.uk}
 }
}

\begin{document}
\maketitle
\begin{abstract}
Content moderation in online platforms is crucial for ensuring activity therein adheres to existing policies, especially as these platforms grow. NLP research in this area has typically focused on automating some part of it given that it is not feasible to monitor all active discussions effectively. Past works have focused on revealing deletion patterns with like sentiment analysis, or on developing platform-specific models such as Wikipedia policy or stance detectors. Unsurprisingly, however, this valuable body of work is rather scattered, with little to no agreement with regards to e.g., the deletion discussions corpora used for training or the number of stance labels. Moreover, while efforts have been made to connect stance with rationales (e.g., to ground a deletion decision on the relevant policy), there is little explanability work beyond that. In this paper, we introduce a suite of experiments on Wikipedia deletion discussions and \texttt{wide-analyis}\footnote{\texttt{wide-analysis} is freely available at \url{https://pypi.org/project/wide-analysis/}} (Wikipedia Deletion Analysis), a Python package aimed at providing ``one-click'' analysis to content moderation discussions. We release all assets associated with \texttt{wide-analysis}, including data, models and the Python package, and a HuggingFace space\footnote{\url{https://huggingface.co/spaces/hsuvaskakoty/wide_analysis_space}} with the goal to accelerate research on automating content moderation in Wikipedia and beyond. A video showcasing  our work is available alongside an example notebook at \url{https://www.youtube.com/watch?v=ILKpKGFgkm8}.
\end{abstract}

\section{Introduction}

Content moderation is often described through the lens of group coordination and communication frameworks \cite{chidambaram2005out,jensen2005collaboration,butler2008don}. From analyzing Facebook moderation in relation with existing policies \cite{sablosky2021dangerous} to tensions between user vs developer-oriented Python mailing lists \cite{barcellini2008user}, or linking deletion decisions to negative sentiment, insults or defamation \cite{risch2018delete,xiao2018sentiments}, NLP plays an important role in automating content moderation activities, especially in large platforms where it is not feasible to keep track of all discussions on content quality simultaneously.

\begin{figure}[!t]
    \centering
    \includegraphics[width=0.5\textwidth]{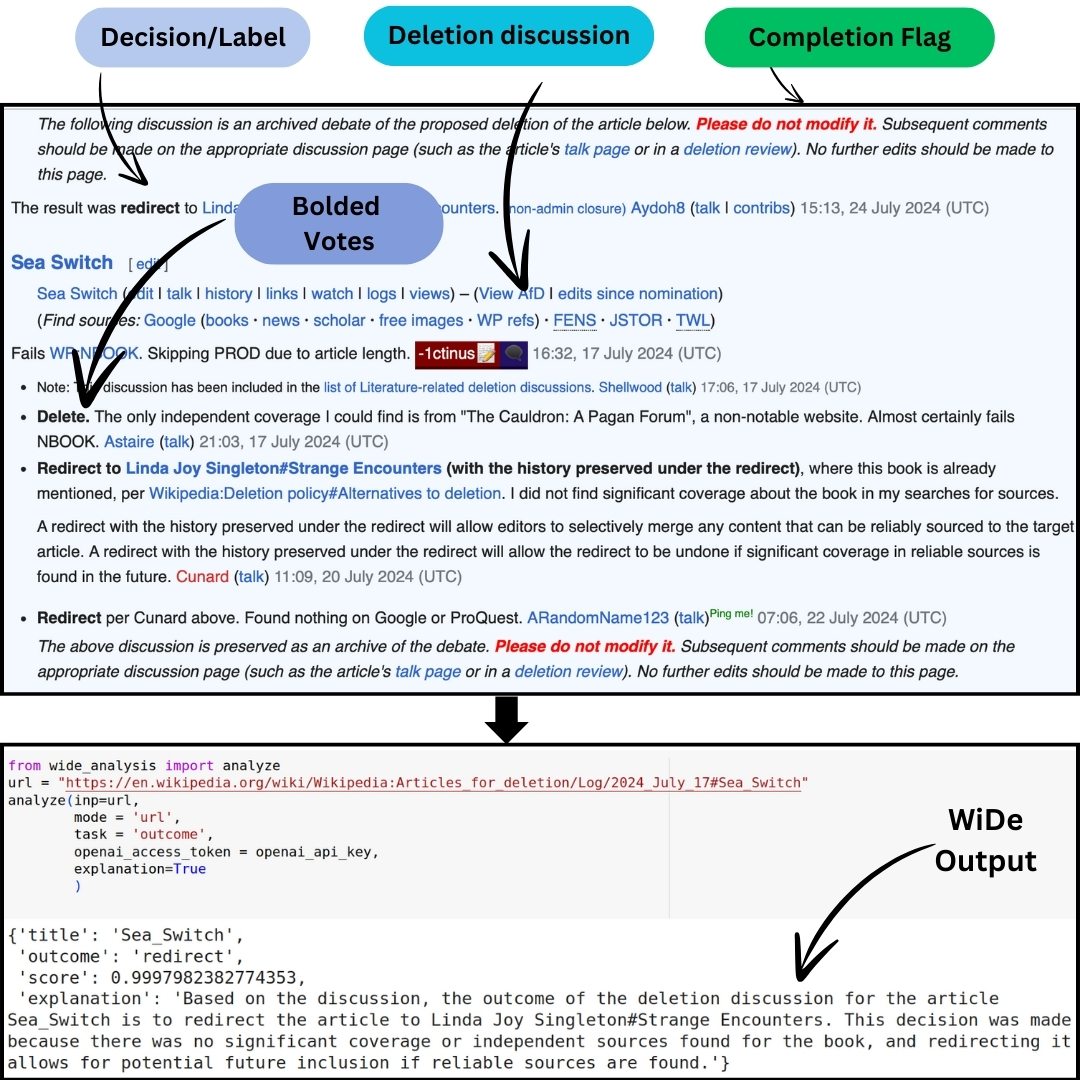}
    \caption{A Wikipedia discussion on an \textit{Article for Deletion} (AfD) and an examples showing how the \texttt{wide-analysis} package enables one-click data retrieval and analysis.}
    \label{fig:example}
\end{figure}

In the specific context of Wikipedia, \citet{mayfield2019stance} investigate different methods for predicting the outcome of a deletion discussion. They survey how outcome decisions (labels) have been historically considered (shedding light on the imbalance of the task, with \texttt{delete} often dominating, ranging from 55 to 64\% of existing datasets, and \texttt{redirect} sometimes accounting only for 2\%). Moreover, they report experiments on classifying discussions outcomes, modeling each of them as a sequence of contributions, with each contribution, in turn, being encoded as an embedding \cite{pennington2014glove,bojanowski2017enriching,devlin2018bert} capturing information such as deletion labels, timestamps and rationale texts. These rationales are nominating statements written in natural language, e.g., ``\emph{Delete. Just a junk article, not notable.}''. These technical (NLP-centric) experiments were then further extended in \citet{mayfield2019analyzing} to explore the \textit{outputs} of, e.g., outcome prediction models. Some of these analysis include ``herding effects'', i.e., measuring the impact early contributors have over the final decision, with the finding that indeed early votes are strong predictors for final outcomes. They also investigate extensively the Wikipedia notability policies, and among others, find that the most successful notability policies are where ``enthusiasts wrote these policies to clearly define notability for an area where the average editor may not know inclusion criteria'', such as astronomy, geographic landmarks or local high schools. Finally, and perhaps contrary to expectation, they also find that active voters do not have significant influence over ``single-contribution'' voters. Another line of work, also based on the data released by \citet{mayfield2019stance}, is concerned with anonymity and persuasiveness, where the goal is to estimate how correlated the anonymity of a user is with their degree of persuasiveness. The results in \citet{xiao2020effects} suggest that participants who are more identifiable tend to be more persuasive in the deletion discussions, and also, that the more identifiable a user is, the more likely they are to invest effort in the discussion (e.g., by producing longer comments and sentences).

Wikipedia policies, in the form of rationales for stance towards a given article, were further studied in \citet{kaffee2023should}. They consider two intertwined tasks. First, stance detection, i.e., given a comment in a Wikipedia deletion discussion, predict its stance towards the discussion in question using as label one of the Wikipedia deletion labels. They encode each instance by concatenating the deletion title (e.g., ``\emph{Deletion of Beta Kappa Gamma}'') and a comment (e.g., ``\emph{Blatant advertising. Fails. The whole article reads like advertisement}'') using the \texttt{[SEP]} token, and predict from a set of labels such as \emph{keep}, \emph{merge}, \emph{delete}, etc. Second, associated policy prediction (which aids in making the first task more transparent), which predicts an existing Wikipedia policy (e.g., \emph{Notability} or \emph{Civility}) from a comment. They experiment with single vs multitask setups, with no clear gains by attempting both tasks simultaneously.

As we can see from this survey, there is significant work on analyzing and further automating deletion discussions in online platforms, with Wikipedia having garnered significant attention. However, the community is currently lacking a centralized toolkit that is able to easily perform tasks such as data gathering and preprocessing, and discussion analysis at different granularities (comment-level and discussion-level), with content analysis stemming from general-purpose systems (e.g., sentiment or offensive language detection) as well as Wikipedia-specific content analyzers (policy prediction or stance detection). Moreover, there is little work on harnessing LLMs for these tasks, either for better classification performance or for summarizing and explaining decisions of text classifiers. We posit that such a centralized tool would significantly accelerate research in Wikipedia content moderation, analogous to what other unification/reconciliation works have achieved in NLP tasks such as lexical semantics \cite{navigli2012babelnet} or Twitter analysis \cite{barbieri2020tweeteval,antypas2023supertweeteval}. 

In this paper, thus, we first present a set of classification experiments on Wikipedia deletion discussions (Section \ref{sec:nlp-tasks}), namely discussion-level fine-grained outcome prediction, and comment-level stance and policy prediction. Then, we introduce the \texttt{wide-analysis} Python package (Section \ref{sec:wide-analyze}), which has two main functionalities: (1) building a cleaned Wikipedia deletion discussion dataset of arbitrary size, given a specific time range or a Wikipedia URL; and (2) enabling ``one-click'' analysis on Wikipedia discussions, which has applications on text classification, argument mining or enabling sociodemographic analysis of online forums. 

\section{NLP on \textit{Articles for Deletion}}
\label{sec:nlp-tasks}

Our overview of related works revealed that, out of the dedicated tasks in Wikipedia's \textit{Articles for Deletion} (AfDs), the most prominent are \textit{outcome} and \textit{stance} detection (concerned with predicting a deletion decision from a full discussion or a single comment, respectively),  and \textit{policy} prediction (which also is typically framed at the comment level). We therefore present a novel set of outcome prediction experiments on a recent snapshot of AfDs (those comprising between January 1st, 2023 to July 18th, 2024), and use existing datasets as starting point for providing a broader set of experiments on policy and stance detection (for the latter task, establishing a new state of the art).

\begin{table}[!th]
\Large
\centering
\resizebox{0.98\columnwidth}{!}{
\begin{tabular}{@{}lr|lr|lr@{}}
\toprule
\multicolumn{2}{c|}{Train set}                         & \multicolumn{2}{c|}{Validation set}                           & \multicolumn{2}{c}{Test set}                          \\ \midrule
delete                    & 7,032                       & delete                    & 1,005                       & delete                    & 2,010                      \\
keep                      & 2,117                       & keep                      & 303                        & keep                      & 605                       \\
redirect                  & 1,648                       & redirect                  & 236                        & redirect                  & 471                       \\
no consensus              & 835                        & no consensus              & 120                        & no consensus              & 239                       \\
merge                     & 735                        & merge                     & 106                        & merge                     & 211                       \\
speedy keep               & 306                        & speedy keep               & 44                         & speedy keep               & 88                        \\
speedy delete             & 168                        & speedy delete             & 24                         & speedy delete             & 49                        \\
withdrawn                 & 122                        & withdrawn                 & 18                         & withdrawn                 & 36                        \\ \midrule
total                     & \multicolumn{1}{l|}{12,963} & Total                     & \multicolumn{1}{l|}{1,856}  & total                     & \multicolumn{1}{l}{3,709}  \\ \bottomrule
\end{tabular}
}
\caption{Outcome prediction dataset statistics.}
\label{tab:dataset-stats}
\end{table}

\subsection{Outcome prediction}
\label{sec:outcome}

\begin{figure}[!t]
    \centering
    \begin{tabular}{c}
        \includegraphics[width=\columnwidth]{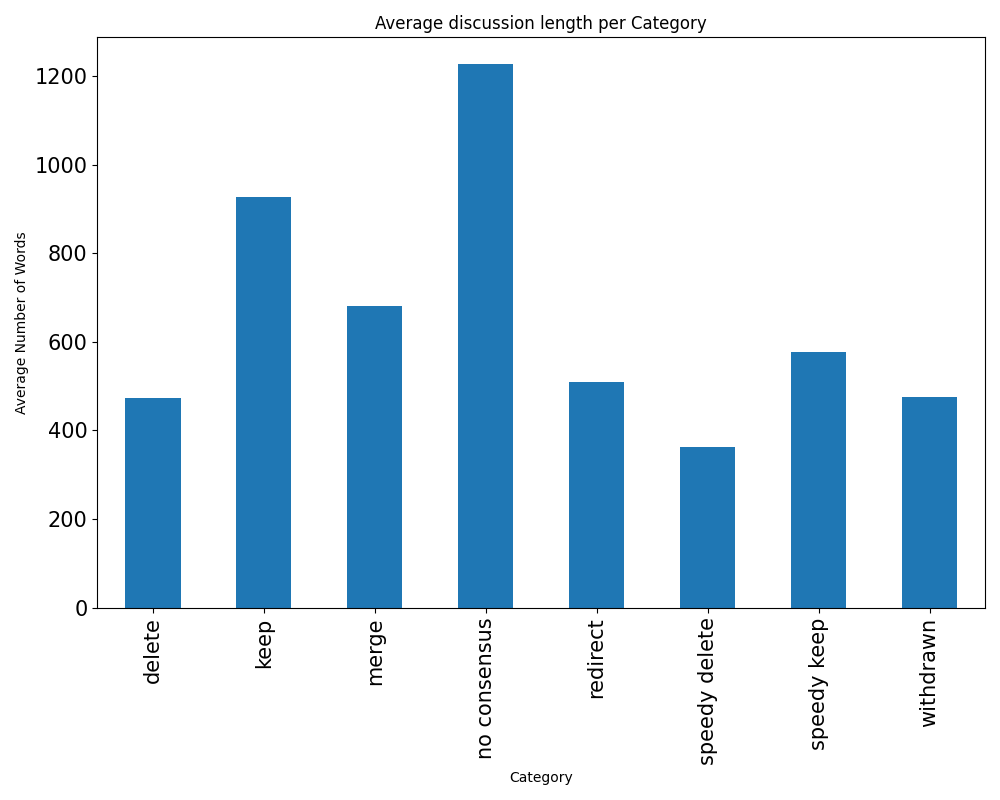} \\
        \text{(a)Average discussion length per category} \\
        \includegraphics[width=\columnwidth]{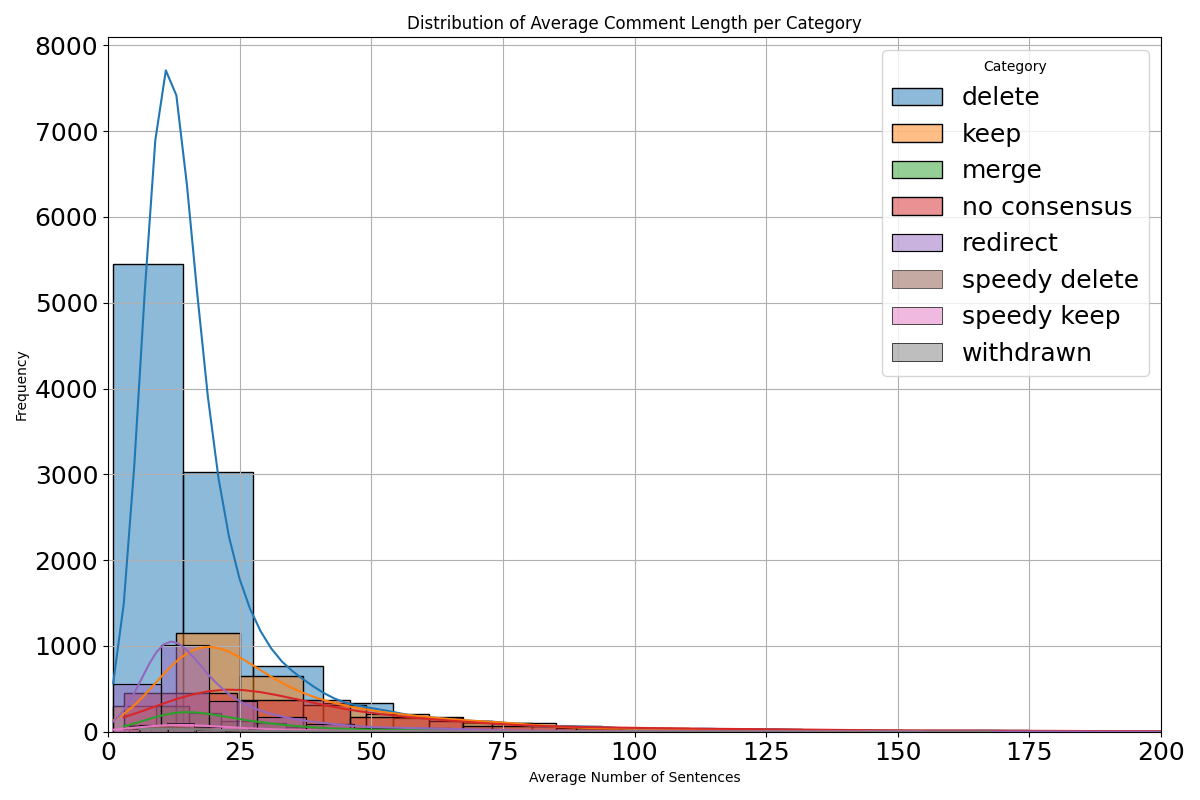} \\
        \text{(b) Average sentence length distribution per category} \\
    \end{tabular}
    \caption{Category-wise discussion length and Sentence length Distribution for Outcome Prediction Dataset}
    \label{fig:dataset-analysis}
\end{figure}

After a Wikipedia article has been nominated for deletion (an AfD), one-week discussions are held, then closed by an administrator with a decision, which often does not deviate from group consensus or lack thereof, in which case the discussion is labeled as \textit{no-consensus} and the article is kept \cite{mayfield2019stance}. In this context, given the discussion occurring around an AfD, \textit{outcome prediction} is concerned with predicting the outcome of that one-week discussion (cf. Figure \ref{fig:example}). In our experiments, we consider the 8 labels listed in Table \ref{tab:dataset-stats}, and evaluate a number of language models, specifically BERT\cite{devlin2018bert} in Base and Large, RoBERTa\cite{liu2019roberta} in Base and Large, DistilBERT\cite{sanh2020distilbert}, and Twitter-RoBERTa-Base\cite{camacho2022tweetnlp}. We also test the following LLMs: GPT-4-omni \cite{achiam2023gpt} and LLama2-7b and LLama3-8b \cite{touvron2023llama,dubey2024llama}. As can be seen from the data statstics from Table \ref{tab:dataset-stats}, \textit{delete} is the most frequent label, covering more than half of the training set, for instance. It is worth noting that despite being similar, there are fundamental differences between some of the labels. For example, \textit{delete} and \textit{keep} differ in that speedy versions constitute a ``limited circumstantial'' approach that can only be considered with respect to some pre-defined criterias (such as \textit{Patent Nonsense} or \textit{Blantant hoax})\footnote{\url{https://en.wikipedia.org/wiki/Wikipedia:Field_guide_to_proper_speedy_deletion}.}, whereas \emph{withdrawn} and \emph{keep} are different in that \textit{keep} is a consensual decision reached by the community, whereas \textit{withdraw} indicates that the nominator of the article deletion has retracted the nomination due to insufficient evidence, but does not necessarily mean that it could not get nominated again in the future. We aim to gain further insights into the average length of comments and discussions per label in this snapshot (cf. Figure \ref{fig:dataset-analysis}), by looking at the length of discussions and the distribution of sentences per category. From the results, it seems clear that \textit{no consensus} debates are typically longer, which is in line with expectation given that these are likely to include contentious points that are hard to resolve (as opposed to straightforward decisions in \textit{speedy delete} or \textit{speedy keep}, for instance). Finally, sentences in \textit{delete}-resulting discussions are rather long, which might suggest preference from editors to engage in longer more authoritative arguments when arguing for deletion.

Concerning the classification experiments themselves, we present results on two setups, one where the deletion is processed \textit{as-is}, which would reflect a real-world use case where we are interested in monitoring likely deletions in currently open discussions without loss of any signal, but also a setup where we mask the ``bolded'' label keywords from the comments, and we learn these classifiers only from the comments, without their explicit stance.

Table \ref{tab:outcome-full} shows classification results for both full and mask setups, with results of LLMs in Full setting shown in Table \ref{tab:llm_perf}. We can unsurprisingly see a drop in performance on the masked setup, although interestingly not as pronounced as one would expect, which suggests that the actual votes can be easily inferred from the comment text themselves. We also note the difference in performance between BERT-based models and LLMs is notable, particularly between RoBERTa-Large and GPT4-omni (few-shot). Both the models are best performing in their setups, but the difference in performance between them is quite notable, which shows the importance of having tailored models for these tasks. Further analysis on low performance of LLMs are to be investigated, which we leave for future scope. The prompt and examples used for the few shot prompts are provided in Appendix \ref{sec:appendix-prompts}.

\begin{table}[!h]
\resizebox{\columnwidth}{!}{
\begin{tabular}{@{}lrrrr|rrrr@{}}
\toprule
                     & \multicolumn{4}{c}{\textbf{Full}}                                                                                                            & \multicolumn{4}{c}{\textbf{Masked}}                                                                                                          \\ \cmidrule(l){2-9} 
                     & \multicolumn{1}{c}{\textbf{Acc.}} & \multicolumn{1}{c}{\textbf{Prec.}} & \multicolumn{1}{c}{\textbf{Rec.}} & \multicolumn{1}{c|}{\textbf{F1}} & \multicolumn{1}{c}{\textbf{Acc,}} & \multicolumn{1}{c}{\textbf{Prec.}} & \multicolumn{1}{c}{\textbf{Rec.}} & \multicolumn{1}{c}{\textbf{F1}} \\ \midrule
RoBERTa Base         & 0.77                              & 0.58                               & 0.55                              & 0.56                            & 0.73                                & 0.60                                 & 0.45                                & 0.49                             \\
RoBERTa Large        & 0.79                              & 0.62                               & 0.55                              & \textbf{0.58}                            & 0.73                                & 0.55                                 & 0.50                                & \textbf{0.52}                             \\
BERT base            & 0.78                              & 0.62                               & 0.54                              & 0.56                            & 0.73                                & 0.54                                 & 0.47                                & 0.49                              \\
BERT Large           & 0.78                              & 0.62                               & 0.56                              & 0.58                            & 0.72                                & 0.56                                & 0.48                                & 0.50                              \\
DistilBERT           & 0.78                              & 0.59                               & 0.53                              & 0.55                            & 0.71                                & 0.64                                 & 0.39                                & 0.43                              \\
Twitter RoBERTa Base & 0.71                                & 0.50                                 & 0.50                                & 0.49                              & 0.71                             & 0.60                                 & 0.41                                & 0.46                              \\ \midrule
\end{tabular}
}
\caption{Outcome prediction results on the full (including bolded votes) setup. We consider BERT-based models, including a Twitter-pretrained model.}
\label{tab:outcome-full}
\end{table}

\begin{table}[]
\resizebox{\columnwidth}{!}{%
\begin{tabular}{llrrrr}
\hline
\multicolumn{1}{c}{\textbf{Model}}             & \multicolumn{1}{c}{\textbf{Setting}} & \multicolumn{1}{c}{\textbf{Acc.}} & \multicolumn{1}{c}{\textbf{Prec.}} & \multicolumn{1}{c}{\textbf{Rec.}} & \multicolumn{1}{c}{\textbf{F1}} \\ \hline
\multicolumn{1}{c}{\multirow{2}{*}{LLama3-8b}} & Zero-shot                            & 0.75                              & 0.40                               & 0.31                              & 0.32                            \\
\multicolumn{1}{c}{}                           & Few-shot                             & 0.75                              & 0.45                               & 0.34                              & 0.35                            \\ \hline
\multirow{2}{*}{LLama2-7b}                     & Zero-shot                            & 0.55                              & 0.35                               & 0.29                              & 0.26                            \\
                                               & Few-shot                             & 0.65                              & 0.45                               & 0.33                              & 0.31                            \\ \hline
\multirow{2}{*}{GPT-4-omni}                    & Zero-shot                            & 0.76                              & 0.49                               & 0.41                              & 0.40                            \\
                                               & Few-shot                             & 0.77                              & 0.57                               & 0.46                              & \textbf{0.45}                   \\ \hline
\end{tabular}%
}
\caption{Performance of outcome prediction task in LLMs for Zero-shot and Few-shot setting.}
\label{tab:llm_perf}
\end{table}
Further insights can be gained by exploring the per-label performance of the best performing model on the full setup (namely, RoBERTa-Large). These are provided in Table \ref{tab:outcome-per-label}, showing that despite frequency being an important factor (in line with previous works), some outcomes seem particularly hard to predict, presumably because they could be easily confused with similar (but operationally different) labels. We present a confusion matrix for this set of results in Appendix \ref{sec:appendix-confusionmatrix}, where we flesh out some of the most significant sources of confusion.

\begin{table}[]
\resizebox{\columnwidth}{!}{%
\begin{tabular}{lrrr|rrr}
\hline
\multirow{2}{*}{\textbf{Label}} & \multicolumn{3}{c|}{\textbf{Full}}                                                                        & \multicolumn{3}{c}{\textbf{Masked}}                                                                      \\ \cline{2-7} 
                                & \multicolumn{1}{c}{\textbf{Prec.}} & \multicolumn{1}{c}{\textbf{Rec.}} & \multicolumn{1}{c|}{\textbf{F1}} & \multicolumn{1}{c}{\textbf{Prec.}} & \multicolumn{1}{c}{\textbf{Rec.}} & \multicolumn{1}{c}{\textbf{F1}} \\ \hline
delete                          & 0.89                               & 0.92                              & 0.90                             & 0.86                                 & 0.89                                & 0.88                              \\
keep                            & 0.70                               & 0.74                              & 0.72                             & 0.64                                 & 0.41                                & 0.51                             \\
merge                           & 0.74                               & 0.72                              & 0.73                             & 0.60                                 & 0.58                                & 0.59                              \\
no consensus                    & 0.37                               & 0.31                              & 0.34                             & 0.34                                 & 0.44                                & 0.39                              \\
speedy keep                     & 0.47                               & 0.41                              & 0.44                             & 0.23                                 & 0.53                                & 0.32                              \\
speedy delete                   & 0.39                               & 0.25                              & 0.31                             & 0.33                                 & 0.22                                & 0.27                              \\
redirect                        & 0.79                               & 0.77                              & 0.78                             & 0.67                                 & 0.62                                & 0.64                              \\
withdrawn                       & 0.60                               & 0.32                              & 0.42                             & 0.35                                 & 0.33                                & 0.34                              \\ \hline
\end{tabular}%
}
\caption{Per-label outcome prediction performance of RoBERTa-Large, the best performing model on the full setup.}
\label{tab:outcome-per-label}
\end{table}

\subsection{Stance and policy prediction}

In this section we report experiments on stance and policy prediction, both of them at the comment level. Since there are established datasets in the literature, in an effort to act as a landing point of research in AfD analysis, we take the dataset introduced in \citet{kaffee2023should} as a starting point. Our stance detection results are comparable to those reported in this paper, whereas for policy prediction, we reduce the number of labels from 92 (in their original dataset) to 15, to make error analysis more feasible and remove a very large number of policy labels which had a negligible number of instances. We find that policies (with instances) like Wikipedia:Wikipuffery (111), Wikipedia:I wouldn't know him from a hole in the ground (109), Wikipedia:Userfication (106), Wikipedia:Record charts (105), Wikipedia:Attack page (102) have a remarkably low number of instances, for the dataset with 437,770 instances. Table \ref{tab:stance-policy} shows results for both tasks. We can see that we are able to push the state of the art of Stance detection up to 0.83 macro F1, from the originally reported 0.80 in \citep{kaffee2023should}. Moreover, in the 15-label setup for policy prediction, we find that the best performing model is RoBERTa-Base with Macro-F1 score 0.67. 


\begin{table}[!h]
\Large
\centering
\resizebox{\columnwidth}{!}{
\begin{tabular}{lrrrrrrrr}
\hline
                     & \multicolumn{4}{c}{\textbf{Stance}}                                                                                                           & \multicolumn{4}{c}{\textbf{Policy}}                                                                                                          \\ \cline{2-9} 
                     & \multicolumn{1}{c}{\textbf{Acc,}} & \multicolumn{1}{c}{\textbf{Prec.}} & \multicolumn{1}{c}{\textbf{Rec.}} & \multicolumn{1}{c|}{\textbf{F1}} & \multicolumn{1}{c}{\textbf{Acc,}} & \multicolumn{1}{c}{\textbf{Prec.}} & \multicolumn{1}{c}{\textbf{Rec.}} & \multicolumn{1}{c}{\textbf{F1}} \\ \hline
RoBERTa Base         & 0.90                               & 0.83                                 & 0.78                               & \multicolumn{1}{r|}{0.81}          & 0.83                                & 0.70                                 &0.56                                & 0.61                              \\
RoBERTa Large        & 0.94                                & 0.85                                 & 0.81                                & \multicolumn{1}{r|}{\textbf{0.83}}          & 0.86                                & 0.74                                 & 0.62                                & \textbf{0.67}                              \\
BERT base            & 0.89                                & 0.81                                 & 0.80                                & \multicolumn{1}{r|}{0.80}          & 0.81                                & 0.65                                 & 0.49                                & 0.55                              \\
BERT Large           & 0.91                                & 0.84                                 & 0.82                                & \multicolumn{1}{r|}{\textbf{0.83}}          & 0.84                                & 0.71                                 & 0.59                                & 0.63                              \\
DistilBERT           & 0.90                                & 0.83                                 & 0.74                                & \multicolumn{1}{r|}{0.78}          & 0.80                                & 0.69                                 & 0.46           & 0.50                              \\
Twitter RoBERTa Base & 0.88                                & 0.80                                & 0.66                                & \multicolumn{1}{r|}{0.70}          & 0.81                                & 0.68                                 & 0.53                                & 0.58                              \\ \hline
\end{tabular}
}
\caption{Stance and policy prediction results.}
\label{tab:stance-policy}
\end{table}

\section{The \texttt{WiDe-analysis} Python package}
\label{sec:wide-analyze}

In this section we introduce the \texttt{WiDe-analysis} Python library, which provides an interface for gathering deletion discussions on demand, as well easy text analysis. We argue this is a helpful tool for unifying and centralizing research on AfDs, and for accelerating research on content moderation more broadly.

\subsection{Gathering and analyzing the data}

It is possible to gather data either providing as input a URL of an currently open discussion, a specific date or a date range, in which case the library will fetch the HTML version of each date within that date range, including start and end. We parse the HTML to extract deletion discussion of each article, and use the specific tags defined by Wikipedia markup to collect the Title, Discussion, Label and Confirmation. We then check if the label adheres to the 8 labels defined (we found that in many cases it can be as minor as uppercase/lowercase or any syntactic issues, i.e., label `withdraw' given as `Withdraw' or `Withdrawn'). We also remove additional texts and markups to keep only the discussion text. Finally, we return the dataframe that contains the title, deletion discussion, and label. For example, in order to gather the dataset used in the outcome prediction experiments, we could simply execute the snippet below:

\begin{mypython}[label=SO-test]
from wide_analysis import data_collect
data_wide = data_collect.collect(
    mode = 'wide_2023')
data_range = data_collect.collect(
    mode = 'date_range', 
    start_date='2023-01-01', 
    end_date='2024-07-18')
assert data_wide == data_range

>>> True
\end{mypython}

\subsection{Text analysis models}
\label{sec:textanalysis}

\texttt{Wide-analysis} provides one-click interfaces to pretrained text analysis models as well as dedicated AfD-relevant models such as the ones we have discussed in Section \ref{sec:nlp-tasks}. Simply by passing the data gathering mode and the task of choice, \texttt{WiDe-analysis} runs a pipeline that involves data gathering and preparation (task-specific), prediction and, optionally, calling an LLM for further insights. Let us show small snippets for each of the tasks.

{\noindent\textbf{Outcome prediction}: Given an input Wikipedia URL\footnote{In this example: \url{https://en.wikipedia.org/wiki/Wikipedia:Articles\_for\_deletion/Log/2024\_July\_15\#Raisul_Islam_Ador}.}, we can perform outcome prediction as follows.
\begin{mypython}[label=SO-test]
from wide_analysis import analyze
url = '<WIKIPEDIA URL>'
predictions = analyze(
    inp=url,
    mode='url', 
    task='outcome',
    openai_access_token='<OPENAI KEY>',
    explanation=True)
print(predictions)
>>> {
    'prediction': 'speedy delete', 
    'probability': 0.99, 
    'explanation': 'The article does not 
    establish the notability of 
    the subject. The references are not 
    reliable and the article 
    is not well written.'
    }
\end{mypython} 
}
{\noindent\textbf{Stance and Policy prediction}:
We can also perform similar operation for other dedicated tasks, like stance or policy prediction. These tasks are performed in comment level, therefore we add a pre-processing step of sentence tokenization of the discussion using PySBD\cite{sadvilkarneumann2020pysbd}, performed automatically through our package. We can perform the tasks of stance and policy predictions as follows. }
\begin{mypython}[label=SO-test]
from wide_analysis import analyze
url = <Wikipedia URL>
analyze(inp=url,
        mode = 'url',
        task = 'stance', #or 'policy'
        )
>>> [{'sentence': <tokenized text>, 
'label': <stance>, 
score: <probablity score>}, 
...]
\end{mypython}

{\noindent\textbf{Additional functionalities}:
As an added functionality, we also integrate sentiment analysis using Twitter-RoBERTa-Base-Sentiment model\cite{loureiro2022timelms} and offensive language detection using Twitter-RoBERTa-Base-Offensive model\cite{barbieri2020tweeteval}. This enables further insights, connecting both sets of text analysis tools. For example, consider the sentiment analysis use case.

\begin{mypython}[label=SO-test]
from wide_analysis import analyze
url = '<WIKIPEDIA URL>'
predictions = analyze(
    inp=url,
    mode='url', 
    task='sentiment')
print(predictions)
>>> [
{'sentence': 'None establish his 
Wikipedia:Notability .  ', 
'sentiment': 'negative', 
'score': 0.515991747379303},
{'sentence': 'The first reference 
is almost identical in wording 
to his official web site.  ', 
'sentiment': 'neutral', 
'score': 0.9082792401313782}, 
]
\end{mypython} 
}


\subsection{\texttt{WiDe-Analysis} Hugging Face Space}
We introduce a Hugging Face space as an initial testing ground for developers. We include our analysis tools in the space. The space contains input for the users in form of deletion discussion URL for an article, and after adding the URL, the user can select among the number of tasks that are given. We load our best performing models for the tasks that we train for (i.e., outcome prediction, stance detection and policy prediction) and load the models we provide in our package (for sentiment and offensive language detection). The result of the task is shown along with the text extracted from URL. Screenshot of our space is shown in Appendix \ref{app:wide_hf_space}.

\subsection{\texttt{WiDe}-powered insights into deletion discussions}
\begin{figure}[!t]
    \centering
    \includegraphics[width=0.5\textwidth]{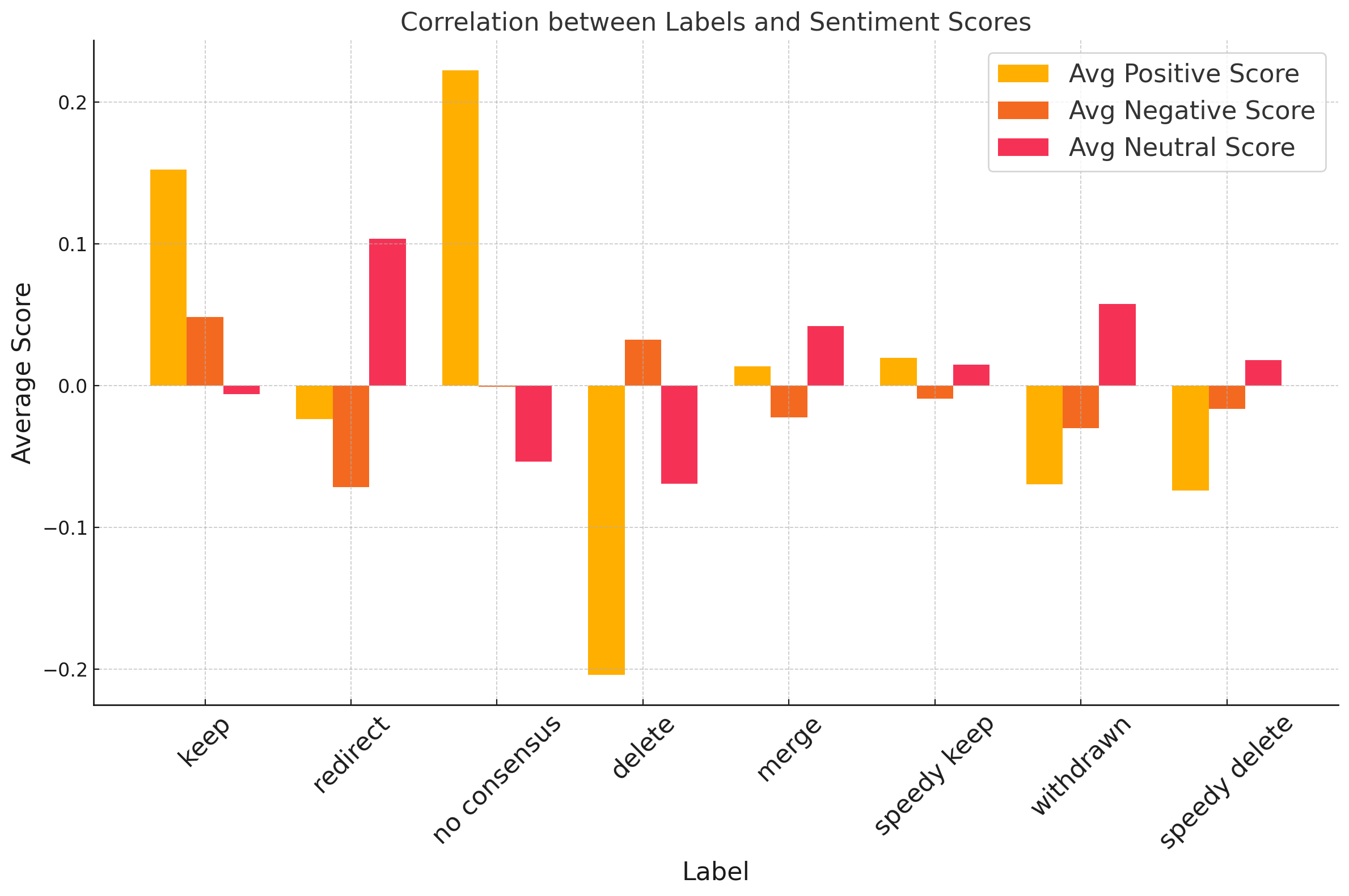}
    \caption{Correlation between sentiments and the labels}
    \label{fig:senti_corr}
\end{figure}

We show the usefulness of the \texttt{WiDe} library by leveraging different functionalities to extract the results of the tasks performed on deletion discussion. We check the correlation between different perspectives derived from text(such as sentiment and stance) with the outcome(label in our dataset).
{\noindent\textbf{Sentiment vs outcome}}: The correlation between sentiments and the outcome labels are shown in Figure \ref{fig:senti_corr}. Although the correlation score is quite low, We find a distinct negative correlation between negative sentiment with `delete', as well as positive sentiment with labels `keep' and `no consensus'.

\begin{figure}[!t]
    \centering
    \includegraphics[width=0.5\textwidth]{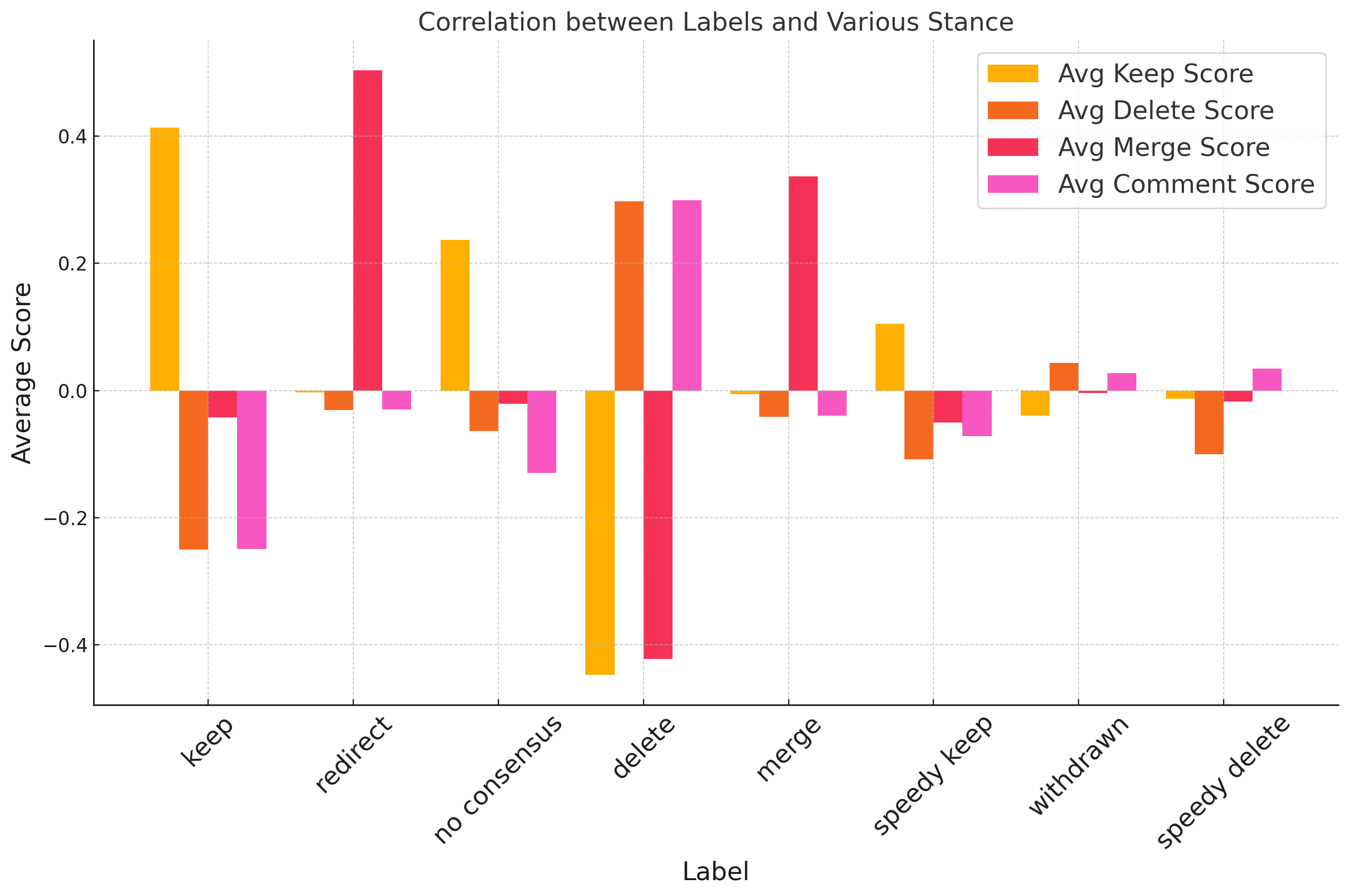}
    \caption{Correlation between stance and the labels}
    \label{fig:stance_corr}
\end{figure}

{\noindent\textbf{Stance vs outcome}}: We use stance detection to extract the stances of the discussions and compare the stances with the outcome labels. The results are shown in Figure \ref{fig:stance_corr}. It's interesting that there's positive correlation between avg merge score and redirect outcome label, and that there's no clear correlation (in fact it's slightly negative) between speedy delete and avg delete scores, which shows that the impact of people advocating for speedy delete on a final delete outcome is not very often.

\section{Conclusion}
\label{sec:conclusions}
We have presented \texttt{WiDe-analysis}, a Python library for performing data collection, preprocessing and text anlaysis on Wikipedia's AfDs. We have also conducted a notable number of experiments on three different tasks, established a new SoTa on stance detection, and provided insights into some of these tasks. Our results have shown that RoBERTa-Large is the most robust model in all the three tasks we train for: outcome prediction, stance detection and policy prediction. We also find that the imbalance in data plays a part in determining the performance of the models (based on the individual results for best performing model), and quite surprisingly LLMs in their in-context learning performs quite poor with respect to the finetuned BERT-family of models in outcome prediction task. We put the analysis of this phenomena as future work and plan to enrich the library with additional functionalities. We make all the assets associated with this paper available to the community: data, code and assets, and we are eager to continue developing assets that contribute to managing large-scale collaborative efforts like Wikipedia. 


\section{Limitations}
\label{sec:limitations}
Our work does not extensively explore all deletion discussions obtainable from Wikipedia(throughout the years), even though it can be obtained using our package. We also do not explore any other LMs except BERT-family of models and smaller versions of LLMs, and due to lack of domain data for sentiment analysis and offensive language detection, we do not train our own models for those tasks. Finally, the tool we propose here can be made better with integration of more analytical tasks and capabilities of model based activities, such as fine-tuning.
\section{Ethics statement}
\label{sec:ethics}
We believe that enhancing quality control for Wikipedia, which is the most popular online encyclopedia through content moderation is always of utmost importance. There is importance of Wikipedia as a viable knowledge source for users, and a data source for today's NLP research is undeniable. This calls for the necessity of tools that enable automated content moderation, so that the discussions that happens behind the curtain of Wikipedia articles regarding its reliability should maintain its standard, while providing resolution for the disputed ones. 

\bibliography{custom}

\appendix

\section{Confounding Factors in Outcome Prediction}
\label{sec:appendix-confusionmatrix}
\begin{figure}[!t]
    \centering
    \begin{tabular}{c}
        \includegraphics[width=\columnwidth]{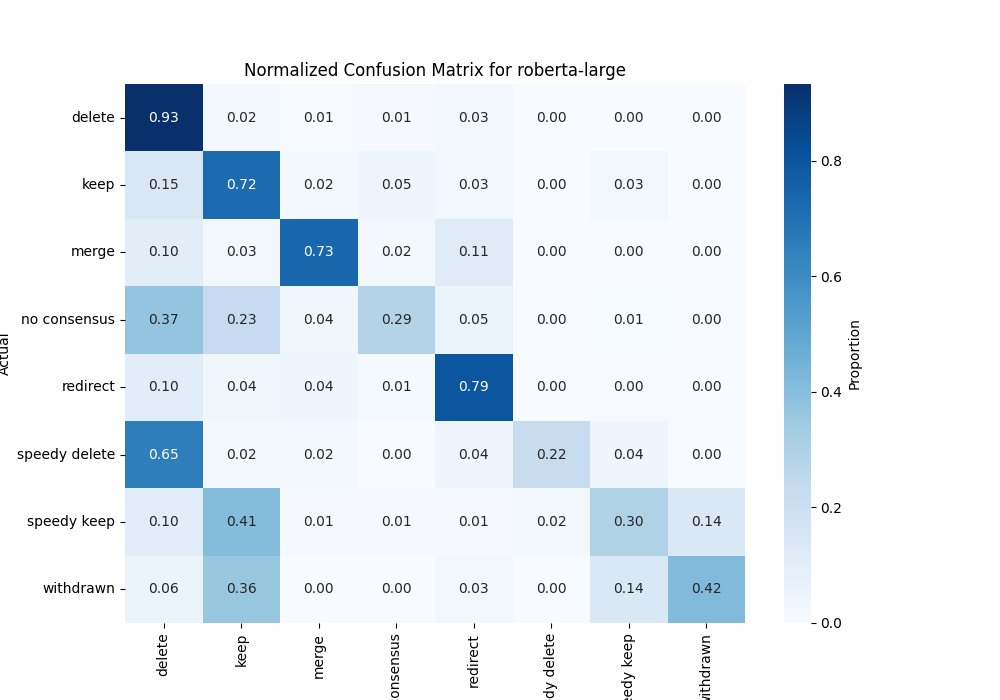} \\
        \text{(a) Full Setting} \\
        \includegraphics[width=\columnwidth]{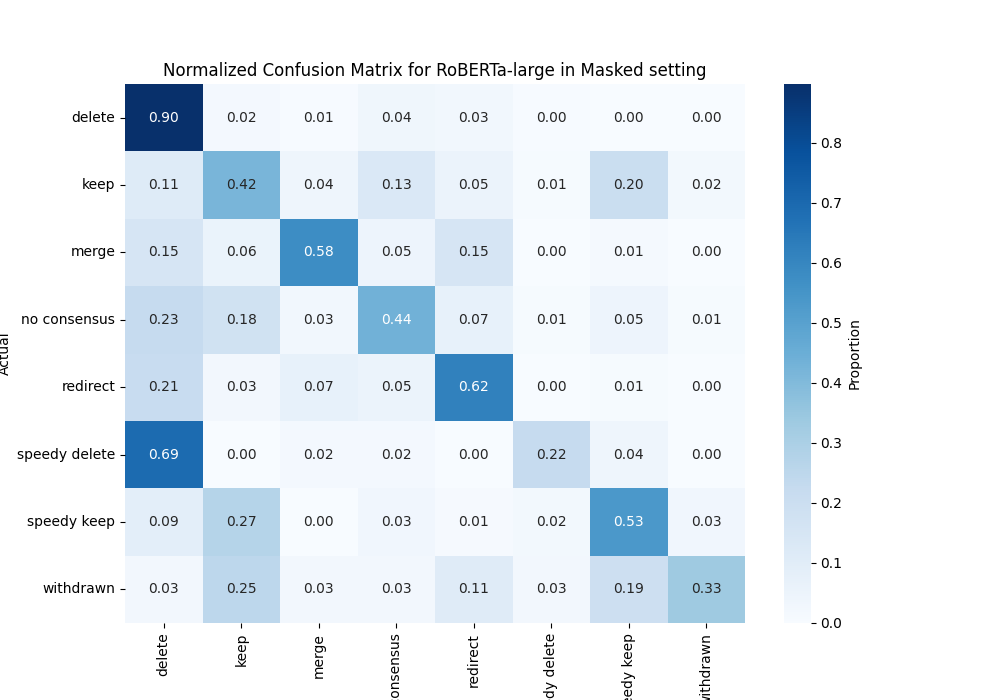} \\
        \text{(b) Masked Setting} \\
    \end{tabular}
    \caption{Confusion Matrix for RoBERTa-Large model in Outcome Prediction Task}
    \label{fig:outcome_roberta}
\end{figure}

The confusion matrix for the best performing model in Outcome Prediction task is shown in Figure \ref{fig:outcome_roberta}, where (a) shows the confusion matrix for Full setting and (b) shows the matrix for masked setting.

\section{\texttt{WiDe-Analysis} Huggingface space}
\label{app:wide_hf_space}
\begin{figure}[!ht]
    \begin{center}
        \includegraphics[width=\columnwidth]{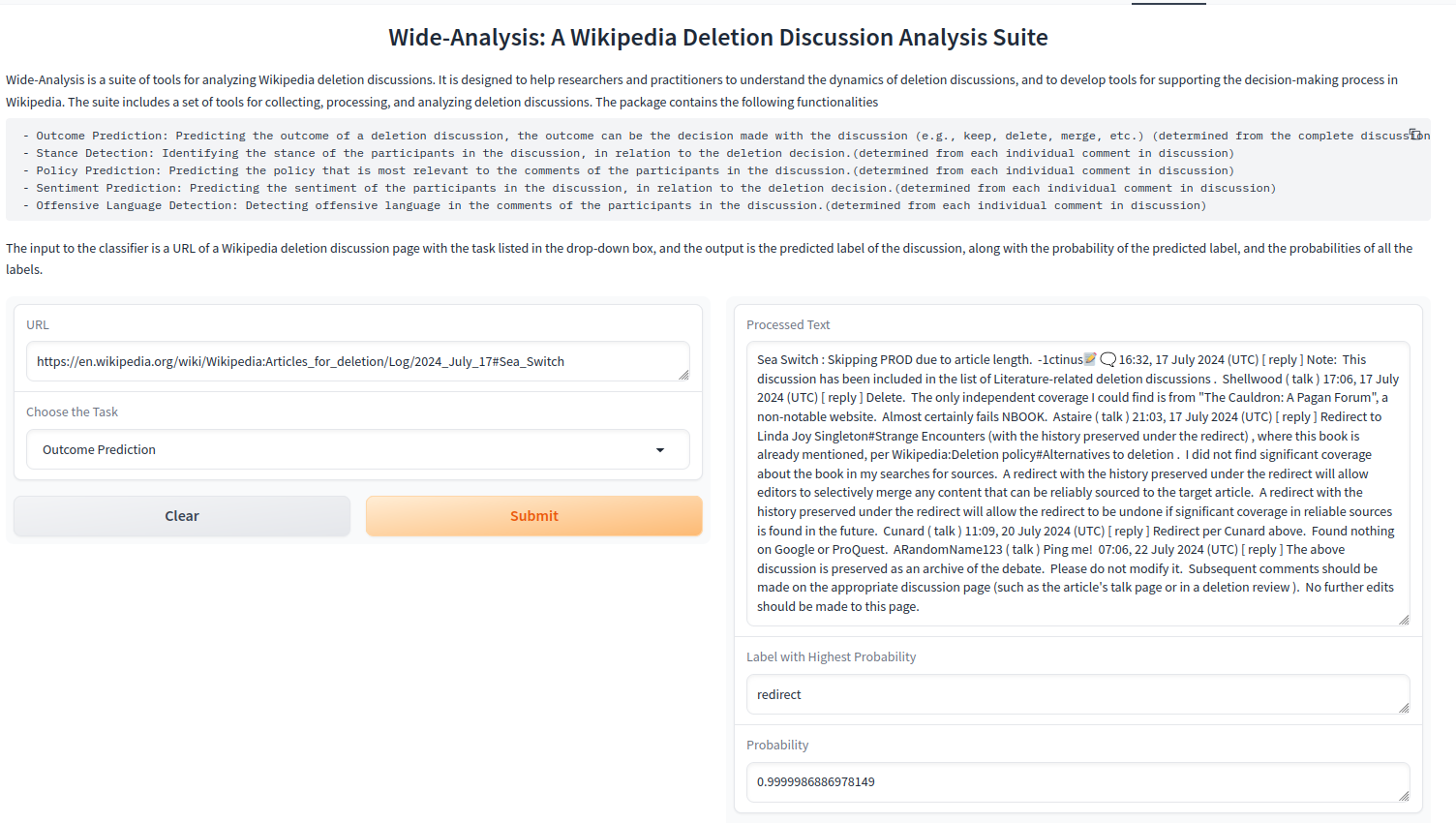}
        \caption{A screenshot of \texttt{WiDe analysis} HuggingFace space.}
        \label{fig:wide_hf}  
    \end{center}
    
\end{figure}

A screenshot of Hugging Face Space showcasing the capabilities of \texttt{WiDe analysis} library is shown in Figure \ref{fig:wide_hf}. 

\newpage
\section{Prompts for Outcome prediction}
\label{sec:appendix-prompts}
\begin{tcolorbox}[width=\textwidth]
\texttt{You are a helpful knowledge management expert, and you excel at identifying the resolution of the Wikipedia deletion discussion for an Article.}

\texttt{Given an article flagged for deletion on Wikipedia along with its deletion discussions, your task is to analyze the article text and discussions to identify the most suitable consensus label based on the deletion discussion.}

\texttt{Your output should be a JSON dictionary with the label that you found and a three-sentence explanation of choosing that label. It is crucial to provide specific reasons based on the content of the deletion discussions and article text. Here is the list of labels with what they mean:}
\begin{itemize}
    \item \texttt{"keep": The article should be kept as it is.}
    \item \texttt{"delete": The article should be deleted.}
    \item \texttt{"merge": The article should be merged with another article.}
    \item \texttt{"redirect": The article should be redirected to another existing article that is a better target for the content.}
    \item \texttt{"withdraw": The nominator withdraws their nomination for deletion.}
    \item \texttt{"no consensus": When there is no clear agreement on the deletion discussion.}
    \item \texttt{"speedy keep": The article should be kept and there are reasons to bypass deletion discussions to keep the article immediately.}
    \item \texttt{"speedy delete": The article should be deleted and there are reasons to bypass deletion discussions to delete the article immediately.}
\end{itemize}
\texttt{Your input will be in the following format:}

\texttt{INPUT:}
\begin{verbatim}
{
     Title: <Article Title>,
     Discussion: <Discussion text>
}
\end{verbatim}

\texttt{OUTPUT:}
\begin{verbatim}
{
    Label: <One of the labels from the list of labels.>,
    Explanation: <Your explanation for the label.>
}
\end{verbatim}
\texttt{Now, you must read the following Input which is a dictionary with Title and deletion discussion. Your task is to analyze the article text and discussions to identify the most suitable consensus label based on the deletion discussion.}
\texttt{INPUT:}
\begin{verbatim}
{
    Title: TOREPLACE_ARTICLE,
    Discussion: TOREPLACE_DISCUSSION
}
\end{verbatim}
\texttt{OUTPUT:}
\end{tcolorbox}

\end{document}